\documentclass[10pt,twocolumn,letterpaper]{article}

\usepackage{cvpr}
\usepackage{times}
\usepackage{epsfig}
\usepackage{graphicx}
\usepackage{amsmath}
\usepackage{amssymb}
\usepackage{multirow}

\makeatletter
\newcommand{\printfnsymbol}[1]{%
  \textsuperscript{\@fnsymbol{#1}}%
}
\makeatother

\usepackage{url}
\usepackage[breaklinks=true,bookmarks=false]{hyperref}
\hypersetup{
    colorlinks=true,
    linkcolor=blue,
    filecolor=blue,      
    urlcolor=magenta,
    citecolor=blue,
}

\cvprfinalcopy % *** Uncomment this line for the final submission

\def\cvprPaperID{2428} % *** Enter the CVPR Paper ID here

\ifcvprfinal\fi
\begin{document}

%%%%%%%%% TITLE
\title{Referring Image Segmentation via Cross-Modal Progressive Comprehension}

\author{Shaofei Huang\textsuperscript{\rm 1,2}\thanks{Equal contribution} \quad Tianrui Hui\textsuperscript{\rm 1,2}\printfnsymbol{1} \quad Si Liu\textsuperscript{\rm 3}\thanks{Corresponding author} \quad Guanbin Li\textsuperscript{\rm 4} \quad Yunchao Wei\textsuperscript{\rm 5}\\ \quad Jizhong Han\textsuperscript{\rm 1,2} \quad Luoqi Liu\textsuperscript{\rm 6} \quad Bo Li\textsuperscript{\rm 3}\\
\textsuperscript{\rm 1} Institute of Information Engineering, Chinese Academy of Sciences\\
\textsuperscript{\rm 2} School of Cyber Security, University of Chinese Academy of Sciences\\
\textsuperscript{\rm 3} School of Computer Science and Engineering, Beihang University\\
\textsuperscript{\rm 4} Sun Yat-sen University
\quad\textsuperscript{\rm 5} University of Technology Sydney
\quad\textsuperscript{\rm 6} 360 AI Institute
}

\maketitle
\thispagestyle{empty}
\pagestyle{empty}

%%%%%%%%% ABSTRACT
\begin{abstract}
   Referring image segmentation aims at segmenting the foreground masks 
   of the entities that can well match the description given in the 
   natural language expression. Previous approaches tackle this problem 
   using implicit feature interaction and fusion between visual and 
   linguistic modalities, but usually fail to explore informative words 
   of the expression to well align features from the two modalities 
   for accurately identifying the referred entity. In this paper, we propose 
   a Cross-Modal Progressive Comprehension (CMPC) module and a Text-Guided 
   Feature Exchange (TGFE) module to effectively address the challenging task. 
   Concretely, the CMPC module first employs entity and attribute words to 
   perceive all the related entities that might be considered by the expression. 
   Then, the relational words are adopted to highlight the correct entity 
   as well as suppress other irrelevant ones by multimodal graph reasoning. 
   In addition to the CMPC module, we further leverage a simple yet effective 
   TGFE module to integrate the reasoned multimodal features from different 
   levels with the guidance of textual information. In this way, 
   features from multi-levels could communicate with each other and be refined 
   based on the textual context. We conduct extensive experiments on four popular 
   referring segmentation benchmarks and achieve new state-of-the-art performances. 
   Code is available at \url{https://github.com/spyflying/CMPC-Refseg}.
\end{abstract}

%%%%%%%%% BODY TEXT
\section{Introduction}

\begin{figure}[t]
   \begin{center}
      \includegraphics[width=0.9\linewidth]{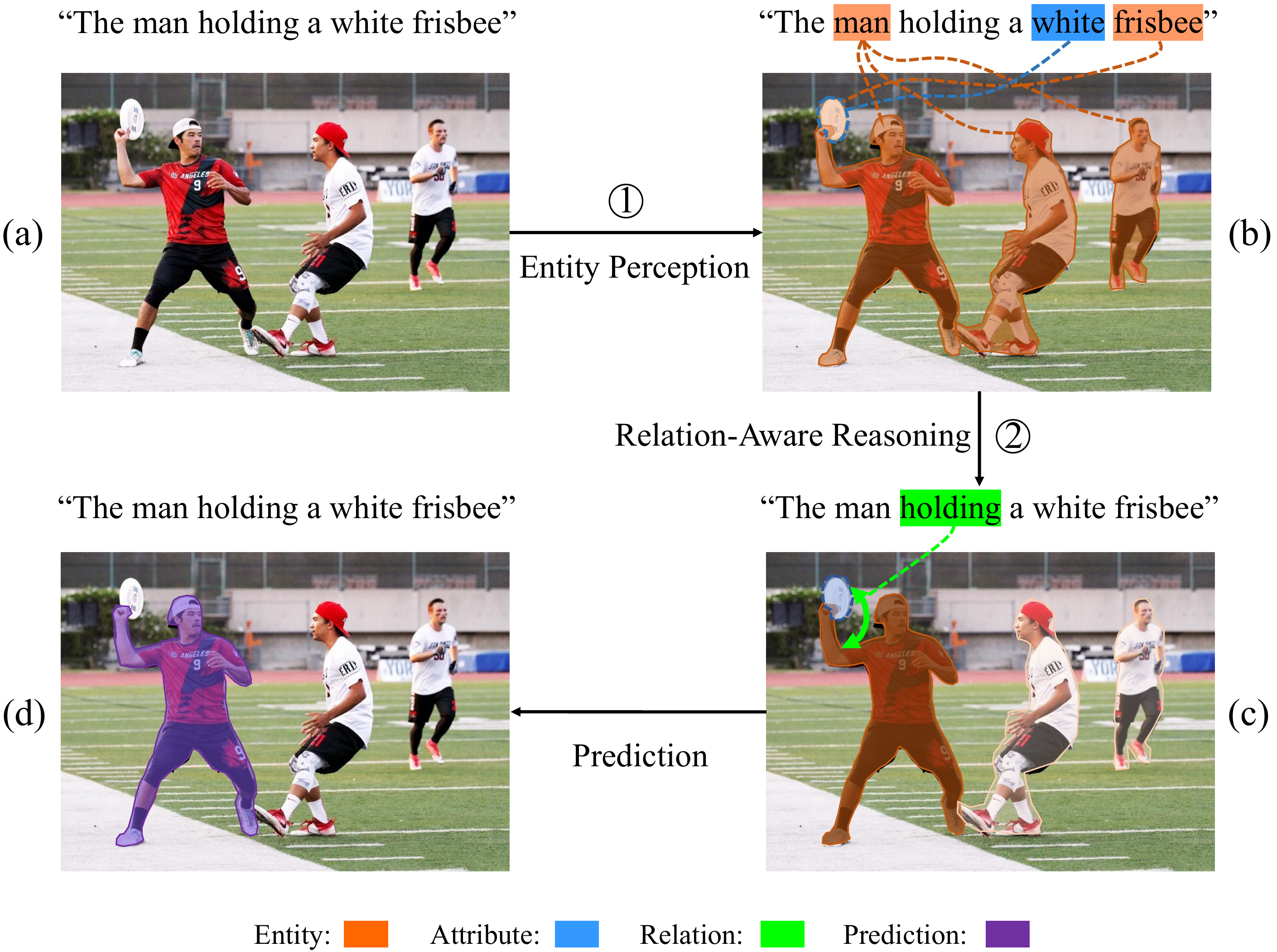}
   \end{center}
      \caption{Interpretation of our progressive referring segmentation method. (a) Input referring 
      expression and image. (b) The model first perceives all the entities 
      described in the expression based on entity words and attribute words, 
      e.g., ``man'' and ``white frisbee'' (orange masks and blue outline). (c) After finding out all the 
      candidate entities that may match with input expression, relational word 
      ``holding'' can be further exploited to highlight the entity involved with the 
      relationship (green arrow) and suppress the others which are not involved. 
      (d) Benefiting from the relation-aware reasoning process, the referred entity is found 
      as the final prediction (purple mask). (Best viewed in color).}
   \label{fig:intro}
\end{figure}

As deep models have made significant progresses in vision or language tasks~\cite{liu2020parsing}\cite{liao2019gps}\cite{jiang2019psgan}\cite{gao2019adversarialnas}\cite{vaswani2017attention}, fields combining them~\cite{ren2020scene}\cite{liao2019ppdm}\cite{zheng2019reasoning} have drawn great attention of researchers. 
In this paper, we focus on the \textit{referring image segmentation} (RIS) 
problem whose goal is to segment the entities described by a natural 
language expression. Beyond traditional semantic segmentation, RIS is a more challenging problem since the expression can 
refer to \emph{objects} or \emph{stuff} belonging to any category in various language 
forms and contain diverse contents including entities, attributes and 
relationships. As a relatively new topic that is still far from being solved, this 
problem has a wide range of potential applications such as interactive 
image editing, language-based robot controlling, etc.
Early works~\cite{hu2016segmentation}\cite{liu2017recurrent}\cite{margffoy2018dynamic}\cite{li2018referring} 
tackle this problem using a straightforward concatenation-and-convolution 
scheme to fuse visual and linguistic features. 
Later works~\cite{shi2018key}\cite{chen2019see}\cite{ye2019cross} further 
utilize inter-modality attention or self-attention to learn only visual embeddings 
or visual-textual co-embeddings for context modeling. However, these methods still 
lack the ability of exploiting different types of informative words in the 
expression to accurately align visual and linguistic features, which is crucial to 
the comprehension of both expression and image.

As illustrated in Figure~\ref{fig:intro} (a) and (b), 
if the referent, i.e., the entity referred to by the expression, is described by 
``The man holding a white frisbee'', a reasonable solution is to tackle the referring 
problem in a progressive way which can be divided into two stages. First, the model is supposed to 
perceive all the entities described in the expression according to entity words and 
attribute words, e.g., ``man'' and ``white frisbee''. Second, as multiple entities of the 
same category may appear in one image, for example, the three men in Figure~\ref{fig:intro} (b), 
the model needs to further reason relationships among entities to highlight the referent 
and suppress the others that are not matched with the relationship cue given in the expression. 
In Figure~\ref{fig:intro} (c), the word ``holding'' which associates ``man'' with ``white frisbee'' 
powerfully guides the model to focus on the referent who holds a white frisbee rather than the other two men, which assists in making correct prediction in Figure~\ref{fig:intro} (d).

Based on the above motivation, we propose a Cross-Modal Progressive Comprehension (CMPC) 
module which progressively exploits different types of words in the expression to segment the referent 
in a graph-based structure. Concretely, our CMPC module consists of two stages. 
First, linguistic features of entity words and attribute words 
(e.g., ``man'' and ``white frisbee'') extracted from 
the expression are fused with visual features extracted from the image to form 
multimodal features where all the entities considered by the expression are perceived. 
Second, we construct a fully-connected spatial graph where each vertex corresponds to an 
image region and feature of each vertex contains multimodal information of the entity. 
Vertexes require appropriate edges to communicate with each other. Naive edges treating 
all the vertexes equally will introduce abundant information and fail to distinguish the referent from other candidates. 
Therefore, our CMPC module employs relational words (e.g., ``holding'') of the expression as a group of routers 
to build adaptive edges to connect spatial vertexes, i.e., entities, that are involved with 
the relationship described in the expression. Particularly, spatial vertexes (e.g., ``man'') that have 
strong responses to the relational words (e.g., ``holding'') will exchange information with others (e.g., ``frisbee'') that also correlate with the 
relational words. Meanwhile, spatial vertexes that have weak responses to the relational words 
will have less interaction with others. After relation-aware reasoning on the multimodal 
graph, feature of the referent can be highlighted while those of the irrelevant 
entities can be suppressed, which assists in generating accurate segmentation.

As multiple levels of features can complement each other~\cite{li2018referring}\cite{ye2019cross}\cite{chen2019see}, 
we also propose a Text-Guided Feature Exchange (TGFE) module to exploit 
information of multimodal features refined by our CMPC module from different levels. 
For each level of multimodal features, our TGFE module utilizes linguistic features 
as guidance to select useful feature channels from other levels to realize information 
communication. After multiple rounds of communication, multi-level features are further 
fused by ConvLSTM~\cite{xingjian2015convolutional} to comprehensively integrate 
low-level visual details and high-level semantics for precise mask prediction.

Our contributions are summarized as follows: 
(1) We propose a Cross-Modal Progressive Comprehension (CMPC) module which first 
perceives all the entities that are possibly referred by the expression, then 
utilizes relationship cues of the input expression to highlight the referent while 
suppressing other irrelevant ones, yielding discriminative feature representations for the referent. 
(2) We also propose a Text-Guided Feature Exchange (TGFE) module to conduct 
adaptive information communication among multi-level features under the guidance of 
linguistic features, which further enhances feature representations for mask prediction. 
(3) Our method achieves new state-of-the-art results on four referring segmentation benchmarks, 
demonstrating the effectiveness of our model.

\begin{figure*}[t]
   \begin{center}
      \includegraphics[width=0.85\linewidth]{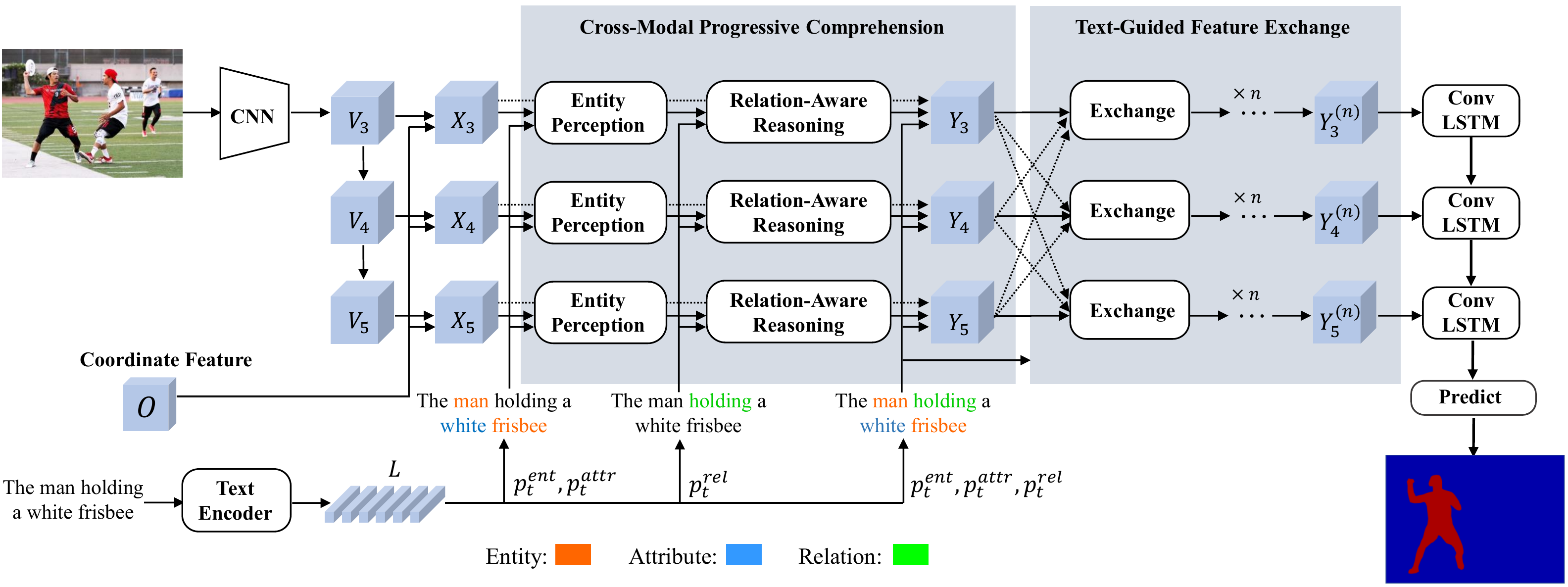}
   \end{center}
      \caption{Overview of our proposed method. Visual features and linguistic features are first progressively aligned by 
      our Cross-Modal Progressive Comprehension (CMPC) module. Then multi-level multimodal features are fed into our Text-Guided Feature Exchange (TGFE) module for information communication across different levels. Finally, multi-level features are fused with ConvLSTM for final prediction.}
   \label{fig:pipeline}
\end{figure*}

%------------------------------------------------------------------------
\vspace{-1mm}
\section{Related Work}
\vspace{-1mm}

\subsection{Semantic Segmentation}
Semantic segmentation has made a huge progress based on 
Fully Convolutional Networks (FCN)~\cite{long2015fully}. 
FCN replaces fully-connected layers in original classification 
networks with convolution layers and becomes the standard 
architecture of the following segmentation methods. 
DeepLab~\cite{chen2014semantic}\cite{chen2017deeplab}\cite{chen2017rethinking} 
introduces atrous convolution with different atrous rates into FCN 
model to enlarge the receptive field of filters and aggregate multi-scale context. 
PSPNet~\cite{zhao2017pyramid} utilizes pyramid pooling operations to extract 
multi-scale context as well. Recent works such as DANet~\cite{fu2019dual} and 
CFNet~\cite{zhang2019co} employ self-attention mechanism~\cite{wang2018non} to 
capture long-range dependencies in deep networks and achieve notable performance. 
In this paper, we tackle the more generalized and challenging semantic segmentation 
problem whose semantic categories are specified by natural language referring expression.

\subsection{Referring Expression Comprehension}
The goal of referring expression comprehension is to localize the entities in the image 
which are matched with the description of a natural language expression. Many works 
conduct localization in bounding box level. 
Liao \textit{et al.}~\cite{liao2019real} performs cross-modality correlation filtering to match multimodal features in real time. 
Relationships between vision and language modalities~\cite{hu2017modeling}\cite{yang2019cross} 
are also modeled to match the expression with most related objects. Modular networks are 
explored in~\cite{yu2018mattnet} to decompose the referring expression into subject, location 
and relationship so that the matching score is more finely computed. 

Beyond bounding box, the referred object can also be localized more precisely with segmentation 
mask. Hu \textit{et al.}~\cite{hu2016segmentation} first proposes the referring segmentation problem and generates 
the segmentation mask by directly concatenating and fusing multimodal features from CNN and 
LSTM~\cite{hochreiter1997long}. In~\cite{liu2017recurrent}, multimodal LSTM is employed to 
sequentially fuse visual and linguistic features in multiple time steps. Based on~\cite{liu2017recurrent}, 
dynamic filters~\cite{margffoy2018dynamic} for each word further enhance multimodal features. 
Fusing multi-level visual features is explored 
in~\cite{li2018referring} to recurrently refine the local details of segmentation mask. As 
context information is critical to segmentation task, Shi \textit{et al.}~\cite{shi2018key} utilizes word 
attention to aggregate only visual context to enhance visual features. 
For multimodal context extraction, cross-modal self-attention is exploited 
in~\cite{ye2019cross} to capture long-range dependencies between each image region and 
each referring word. Visual-textual co-embedding is explored in~\cite{chen2019see} to measure 
compatibility between referring expression and image. 
Adversarial learning~\cite{qiu2019referring} and cycle-consistency~\cite{chen2019referring} 
between referring expression and its reconstructed caption are also investigated to boost 
the segmentation performance. In this paper, we propose to progressively 
highlight the referent via entity perception and relation-aware reasoning for 
accurate referring segmentation.

\subsection{Graph-Based Reasoning}
It has been shown that graph-based models are effective for context reasoning in many tasks. 
Dense CRF~\cite{chandra2017dense} is a widely used graph model for post-processing in image 
segmentation. Recently, Graph Convolution Networks (GCN)~\cite{chandra2017dense} becomes popular 
for its superiority on semi-supervised classification. Wang \textit{et al.}~\cite{wang2018videos} 
construct a spatial-temporal graph using region proposals as vertexes and conduct context reasoning with GCN, 
which performs well on video recognition task. Chen \textit{et al.}~\cite{chen2019graph} propose 
a global reasoning module which projects visual feature into an interactive space and conducts graph 
convolution for global context reasoning. The reasoned global context is projected back to the coordinate space 
to enhance original visual feature. There are several concurrent 
works~\cite{li2018beyond}\cite{liang2018symbolic}\cite{zhang2019latentgnn} sharing the same idea of 
projection and graph reasoning with different implementation details. In this paper, we propose to 
regard image regions as vertexes to build a spatial graph where each vertex saves multimodal feature 
vector as its state. Information flow among vertexes is routed by relational words in the referring 
expression and implemented using graph convolution. After the graph reasoning, image regions can 
generate accurate and coherent responses to the referring expression.

\begin{figure*}[htbp]
   \begin{center}
      \includegraphics[width=0.8\linewidth]{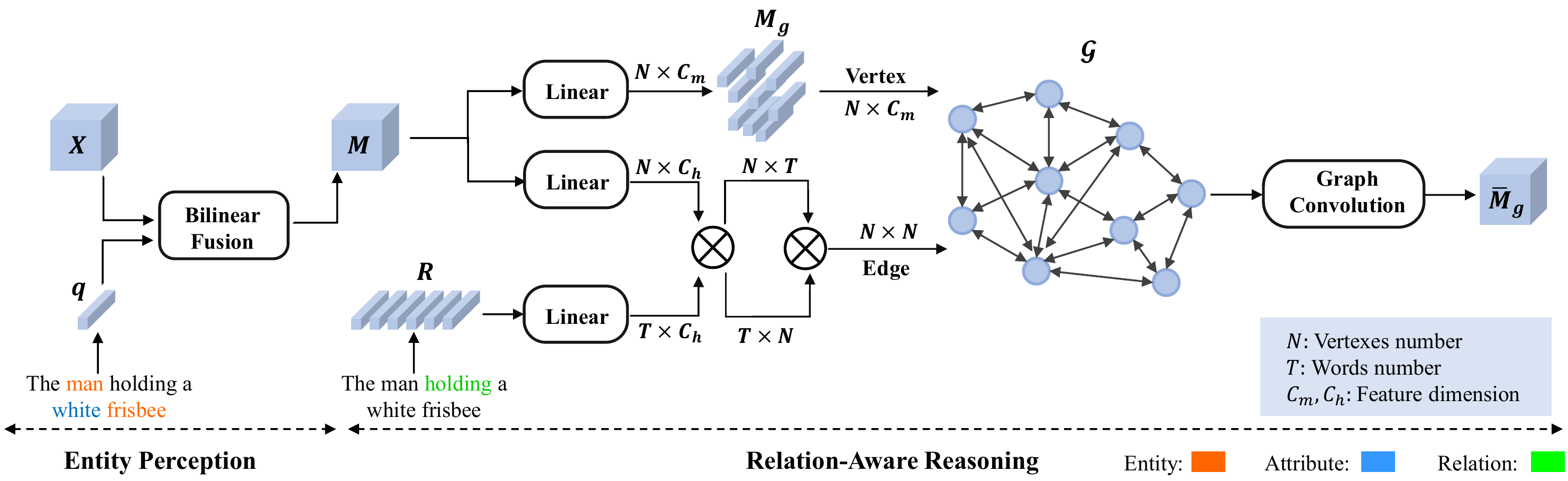}
   \end{center}
      \caption{Illustration of our Cross-Modal Progressive Comprehension module which consists of two stages. 
      First, visual features $X$ are bilinearly fused with linguistic features $q$ of entity words and 
      attribute words for Entity Perception (EP) stage. Second, multimodal features $M$ from EP stage are fed 
      into Relation-Aware Reasoning (RAR) stage for feature enhancement. A multimodal fully-connected graph 
      $\mathcal{G}$ is constructed with each vertex corresponds to an image region on $M$. The adjacency matrix 
      of $\mathcal{G}$ is defined as the product of the matching degrees between vertexes and relational words in 
      the expression. Graph convolution is utilized to reason among vertexes so that the referent could be 
      highlighted during the interaction with correlated vertexes.}
   \label{fig:reason}
\end{figure*}

%------------------------------------------------------------------------
\vspace{-1mm}
\section{Method}
\vspace{-1mm}

Given an image and a natural language expression, the goal of our model is to segment the 
corresponding entity referred to by the expression, i.e., the referent. 
The overall architecture of our model is illustrated in Figure~\ref{fig:pipeline}. 
We first extract the visual features of the image with a CNN backbone and the linguistic features 
of the expression with a text encoder. 
A novel Cross-Modal Progressive Comprehension (CMPC) module is proposed to progressively highlight 
the referent and suppress the others via entity perception and subsequent relation-aware 
reasoning on spatial region graph.
The proposed CMPC module is applied to multiple levels of visual features respectively and the corresponding 
outputs are fed into a Text-Guided Feature Exchange (TGFE) module to communicate information under the guidance of linguistic modality.
After the communication, multi-level features are finally fused with ConvLSTM~\cite{xingjian2015convolutional} 
to make the prediction.
We will elaborate each part of our method in the rest subsections.

\vspace{-1mm}
\subsection{Visual and Linguistic Feature Extraction}
\vspace{-1mm}

As shown in Figure~\ref{fig:pipeline}, our model takes an image and an expression as inputs. 
The multi-level visual features are extracted with a CNN backbone and respectively fused with an $8$-D 
spatial coordinate feature $O \in \mathbb{R} ^ {H \times W \times 8}$ using a $1 \times 1$ convolution 
following prior works~\cite{liu2017recurrent}\cite{ye2019cross}. 
After the convolution, each level of visual features are transformed to the 
same size of $\mathbb{R} ^ {H \times W \times C_v}$, 
with $H$, $W$ and $C_v$ being the height, width and channel dimension of the visual features. 
The transformed visual features are denoted as \{$X_3$, $X_4$, $X_5$\} corresponding to the output of 
the $3$rd, $4$th and $5$th stages of CNN backbone (e.g., ResNet-101~\cite{he2016deep}).
For ease of presentation, we denote a single level of visual features as $X$ in Sec.~\ref{sec:cmpc}. 
The linguistic features $L = \{l_1, l_2, ..., l_T\}$ is extracted with a language 
encoder (e.g., LSTM~\cite{hochreiter1997long}), where $T$ is the length 
of expression and $l_i \in \mathbb{R} ^ {C_l} (i \in \{1, 2, ..., T\})$ denotes 
feature of the $i$-th word.

\subsection{Cross-Modal Progressive Comprehension}
\label{sec:cmpc}
\vspace{-1mm}

As many entities may exist in the image, it is natural to progressively narrow down the candidate set 
from all the entities to the actual referent. 
In this section, we propose a Cross-Modal Progressive Comprehension (CMPC) module which consists of 
two stages, as illustrated in Figure~\ref{fig:reason}. 
The first stage is entity perception. We associate linguistic features of entity words and attribute words 
with the correlated visual features of spatial regions using bilinear fusion~\cite{BenMUTAN} to obtain the multimodal 
features $M\in \mathbb{R} ^ {H \times W \times C_m}$. All the candidate entities are perceived by the fusion. 
The second stage is relation-aware reasoning. A fully-connected multimodal graph is constructed 
over $M$ with relational words serving as a group of routers to connect vertexes.
Each vertex of the graph represents a spatial region on $M$.
By reasoning among vertexes of the multimodal graph, the responses of the referent
 matched with the relationship cue are highlighted while those of non-referred 
ones are suppressed accordingly.
Finally, the enhanced multimodal features $\bar{M_g}$ are further fused with visual and linguistic features.

\textbf{Entity Perception. }Similar to~\cite{yang2019cross}, we classify the words into $4$ types, including entity, attribute, relation and unnecessary word. 
A $4$-D vector is predicted for each word to indicate the probability of it being the four types respectively.
We denote the probability vector for word $t$ as $p_t = [p_{t}^{ent}, p_{t}^{attr}, p_{t}^{rel}, p_{t}^{un}] \in \mathbb{R} ^ 4$ and calculate it as:
\begin{equation}
   \label{eq:prob_pred}
   p_t = softmax(W_2\sigma(W_1l_t+b_1)+b_2),
\end{equation}
where $W_1 \in \mathbb{R} ^ {C_n \times C_l}$, $W_2 \in \mathbb{R} ^ {4 \times C_n}$, $b_1 \in \mathbb{R} ^ {C_n}$ and $b_2 \in \mathbb{R} ^ {4}$ are learnable parameters, $\sigma(\cdot)$ is sigmoid function, 
$p_{t}^{ent}$, $p_{t}^{attr}$, $p_{t}^{rel}$ and $p_{t}^{un}$ denote the probabilities of word $t$ being the entity, attribute, relation and unnecessary word respectively. 
Then the global language context of entities $q \in \mathbb{R} ^ {C_l}$ could be calculated as a weighted combination of the all the words in the expression:
\begin{equation} 
   \label{eq:setence_feat}
   q = \sum\limits_{t=1}^{T}{(p_t^{ent}+p_t^{attr})l_t}.
\end{equation}

Next, we adopt a simplified bilinear fusion strategy~\cite{BenMUTAN} to associate $q$ with the visual feature of each spatial region:
\begin{equation}
   \label{eq:mutan}
   M_i = (qW_{3i})\odot (XW_{4i}),
\end{equation}
\begin{equation}
   \label{eq:ta}
   M = \sum\limits_{i=1}^{r}{M_i}
\end{equation}
where $W_{3i} \in \mathbb{R}^{C_l \times C_m}$ and $W_{4i} \in \mathbb{R}^{C_v \times C_m}$ are learnable parameters, $r$ is a hyper-parameter and $\odot$ denotes element-wise product.
By integrating both visual and linguistic context into the multimodal features, all the entities that might be referred to by the expression are perceived appropriately.

\textbf{Relation-Aware Reasoning. }To selectively highlight the referent, we construct a fully-connected graph over the mutimodal features $M$ and conduct reasoning over the graph according to relational cues in the expression.
Formally, the multimodal graph is defined as $\mathcal{G}  = (\mathcal{V}, \mathcal{E}, M_g, A)$ where $\mathcal{V}$ and $\mathcal{E}$ are the sets of vertexes and edges, $M_g = \{m_i\}_{i=1}^{N} \in \mathbb{R} ^ {N \times C_m}$ is the set of vertex features, $A \in \mathbb{R} ^ {N \times N}$ is the adjacency matrix and $N$ is number of vertexes.

Details of relation-aware reasoning is illustrated in the right part of Figure~\ref{fig:reason}.
As each location on $M$ represents a spatial region on the original image, we regard each region as a vertex of the graph and the multimodal graph is composed of $N = H \times W$ vertexes in total.
After the reshaping operation, a linear layer is applied to $M$ to transform it into the features of vertexes $M_g$. 
The edge weights depend on the affinities between vertexes and relational words in the referring expression. 
Features of relational words $R = \{r_t\}_{t=1}^T \in \mathbb{R} ^{T \times C_l}$ are calculated as:
\begin{equation}
   \label{eq:relation-extract}
   r_t = p_t^{rel}l_t,~~~~t=1, 2, ..., T.
\end{equation}
As shown in Figure~\ref{fig:reason}, adjacency matrix $A$ is formulated as:
\begin{equation}
   \label{eq:inter}
   B = (M_gW_5)(RW_6)^T,
\end{equation}
\begin{equation}
   \label{eq:u1}
   B_1 = softmax(B),
\end{equation}
\begin{equation}
   \label{eq:u2}
   B_2 = softmax(B^T),
\end{equation}
\begin{equation}
   \label{eq:adj_matrx}
   A = B_1B_2,
\end{equation}
where $W_5 \in \mathbb{R} ^{C_m \times C_h}$ and $W_6 \in \mathbb{R} ^{C_l \times C_h}$ are learnable parameters. $B \in \mathbb{R} ^{N \times T}$ is the affinity matrix between $M_g$ and $R$. We apply the softmax function along the second and first dimension of $B$ to obtain $B_1 \in \mathbb{R} ^ {N \times T}$ and $B_2 \in \mathbb{R} ^ {T \times N}$ respectively.
$A$ is obtained by matrix product of $B_1$ and $B_2$.
Each element $A_{ij}$ of $A$ represents the normalized magnitude of information flow from the spatial region $i$ to the region $j$, which depends on their affinities with relational words in the expression.
In this way, relational words of the expression can be leveraged as a group of routers to build adaptive edges connecting vertexes.

After the construction of multimodal graph $\mathcal{G}$, we apply graph convolution~\cite{kipf2016semi} to it as follow:
\begin{equation}
   \label{eq:gcn}
   \bar{M_g} = (A+I)M_gW_7,
\end{equation}
where $W_7 \in \mathbb{R} ^{C_m \times C_m}$ is a learnable weight matrix.
$I$ is identity matrix serving as a shortcut to ease optimization.
The graph convolution reasons among vertexes, i.e., image regions, so that the referent is selectively highlighted according to the relationship cues while other irrelevant ones are suppressed, which assists in generating more discriminative feature representations for referring segmentation.

Afterwards, reshaping operation is applied to obtain the enhanced multimodal 
features $\bar{M_g} \in \mathbb{R} ^ {H \times W \times C_m}$.
To incorporate the textual information, we first combine features of all necessary 
words into a vector $s \in \mathbb{R} ^ {C_l}$ with the predefined probability vectors:
\begin{equation}
   \label{eq:sentence}
   s = \sum_{t=0}^{T}{(p_t^{ent}+p_t^{attr}+p_t^{rel})l_t}.
\end{equation}
We repeat $s$ for $H \times W$ 
times and concatenate it with $X$ and $\bar{M_g}$ along channel dimension following with a $1 \times 1$ convolution to get the output features $Y \in \mathbb{R} ^ {H \times W \times C_m}$, which is equipped with multimodal context for the referent.

\begin{table*}[t]
   \centering
   \begin{tabular}{c|ccc|ccc|c|c}
       \hline
        Method & & UNC & & & UNC+ & & G-Ref & ReferIt \\
        & val & testA & testB & val & testA & testB & val & test \\
       \hline
       LSTM-CNN~\cite{hu2016segmentation} & - & - & - & - & - & - & 28.14 & 48.03 \\
       RMI~\cite{liu2017recurrent} & 45.18 & 45.69 & 45.57 & 29.86 & 30.48 & 29.50 & 34.52 & 58.73 \\
       DMN~\cite{margffoy2018dynamic} & 49.78 & 54.83 & 45.13 & 38.88 & 44.22 & 32.29 & 36.76 & 52.81 \\
       KWA~\cite{shi2018key} & - & - & - & - & - & - & 36.92 & 59.09 \\
       ASGN~\cite{qiu2019referring} & 50.46 & 51.20 & 49.27 & 38.41 & 39.79 & 35.97 & 41.36 & 60.31 \\
       RRN~\cite{li2018referring} & 55.33 & 57.26 & 53.95 & 39.75 & 42.15 & 36.11 & 36.45 & 63.63 \\
       MAttNet~\cite{yu2018mattnet} & 56.51 & 62.37 & 51.70 & 46.67 & 52.39 & 40.08 & n/a & - \\
       CMSA~\cite{ye2019cross} & 58.32 & 60.61 & 55.09 & 43.76 & 47.60 & 37.89 & 39.98 & 63.80 \\
       CAC~\cite{chen2019referring} & 58.90 & 61.77 & 53.81 & - & - & - & 44.32 & - \\
       STEP~\cite{chen2019see} & 60.04 & 63.46 & 57.97 & 48.19 & 52.33 & 40.41 & 46.40 & 64.13 \\
       \hline
       Ours & \textbf{61.36} & \textbf{64.53} & \textbf{59.64} & \textbf{49.56} & \textbf{53.44} & \textbf{43.23} & \textbf{49.05} & \textbf{65.53} \\
       \hline
   \end{tabular}
   \caption{Comparison with state-of-the-art methods on four benchmark datasets using \textit{overall IoU} as metric. ``n/a'' denotes MAttNet does not use the same split as other methods.}
   \label{tab:sota}
\end{table*}

\subsection{Text-Guided Feature Exchange}
As previous works~\cite{li2018referring}\cite{ye2019cross} show that multi-level 
semantics are essential to referring segmentation, we further introduce a Text-Guided 
Feature Exchange (TGFE) module to communicate information among multi-level features 
based on the visual and language context. 
As illustrated in Figure~\ref{fig:pipeline}, the TGFE module takes $Y_3, Y_4, Y_5$ and 
word features $[l_1, l_2, ..., l_T]$ as input. After $n$ rounds of feature exchange, 
$Y_3^{(n)}, Y_4^{(n)}, Y_5^{(n)}$ are produced as outputs.

To get $Y_i^{(k)}, i \in \{3, 4, 5\}, k \geq 1$, we first extract a global vector $g_i^{(k-1)} \in \mathbb{R} ^ {C_m}$ 
of $Y_i^{(k-1)}$ by weighted global pooling:
\begin{equation}
   \label{eq:global_vec}
   g_i^{(k-1)} = \Lambda_i^{(k-1)} Y_i^{(k-1)},
\end{equation}
where the weight matrix $\Lambda_i^{(k-1)} \in \mathbb{R}^{HW}$ is derived from:
\begin{equation}
   \label{eq:pool_weight}
   \Lambda_i^{(k-1)} = (sW_8) (Y_i^{(k-1)}W_9)^T,
\end{equation}
where $W_8 \in \mathbb{R}^{C_l \times C_h}$ and $W_9 \in \mathbb{R}^{C_m \times C_h}$ are transforming matrices.
Then a context vector $c_i^{(k-1)}$ which contains multimodal context of $Y_i^{(k-1)}$ is 
calculated by fusing $s$ and $g_i^{(k-1)}$ with a fully connected layer. 
We finally select information correlated with level $i$ from features 
of other two levels to form the refined features of level $i$ at round $k$:
\begin{equation}
   \label{eq:extract}
   Y_i^{(k)} = 
      \begin{cases}
         Y_i^{(k-1)} + \sum\limits_{j \in \{3, 4, 5\} \backslash \{i\}}{\sigma(c_i^{(k-1)}) \odot Y_j^{(k-1)}}, k \geq 1 \\
         \\
         Y_i, k = 0
      \end{cases}
\end{equation}
where $\sigma(\cdot)$ denotes the sigmoid function.  
After $n$ rounds of feature exchange, features of each level are mutually refined to 
fit the context referred to by the expression.
We further fuse the output features $Y_3^{(n)}$, $Y_4^{(n)}$ and $Y_5^{(n)}$ with 
ConvLSTM~\cite{xingjian2015convolutional} for harvesting the final prediction.

\section{Experiments}
\subsection{Experimental Setup}
\label{sec:setup}
\textbf{Datasets. }We conduct extensive experiments on four benchmark datasets for referring image segmentation including UNC~\cite{yu2016modeling}, UNC+~\cite{yu2016modeling}, G-Ref~\cite{mao2016generation} and ReferIt~\cite{kazemzadeh2014referitgame}.

UNC, UNC+ and G-Ref datasets are all collected based on MS-COCO~\cite{lin2014microsoft}. They contain 
$19,994$, $19,992$ and $26,711$ images with $142,209$, $141,564$ and $104,560$ referring expressions 
for over $50,000$ objects, respectively. UNC+ has no location words and G-Ref contains much longer 
sentences (average length of $8.4$ words) than others (less than $4$ words), making them more challenging 
than UNC dataset. ReferIt dataset is collected on IAPR TC-12~\cite{escalante2010segmented} and contains 
$19,894$ images with $130,525$ expressions for $96,654$ objects (including stuff).

\textbf{Implementation Details. }We adopt DeepLab-101~\cite{chen2017deeplab} pretrained on PASCAL-VOC 
dataset~\cite{everingham2010pascal} as the CNN backbone following prior works~\cite{ye2019cross}\cite{li2018referring} 
and use the output of Res$3$, Res$4$ and Res$5$ for multi-level feature fusion. 
Input images are resized to $320 \times 320$.
Channel dimensions of features are set as $C_v = C_l = C_m = C_h = 1000$ and the cell size of ConvLSTM~\cite{xingjian2015convolutional} is set to $500$.
When comparing with other methods, the hyper-parameter $r$ of bilinear fusion is set to $5$ and the 
number of feature exchange rounds $n$ is set to $3$. GloVe word embeddings~\cite{pennington2014glove} 
pretrained on Common Crawl 840B tokens are adopted following~\cite{chen2019see}. 
Number of graph convolution layers is set to $2$ on G-Ref dataset and $1$ on others.
The network is trained using Adam optimizer~\cite{kingma2014adam} with the initial learning rate of $2.5e^{-4}$ and weight decay of $5e^{-4}$. 
Parameters of CNN backbone are fixed during training. 
The standard binary cross-entropy loss averaged over all pixels is leveraged for training. 
For fair comparison with prior works, DenseCRF~\cite{krahenbuhl2011efficient} is adopted to refine the segmentation masks.

\textbf{Evaluation Metrics. }Following prior works~\cite{hu2016segmentation}\cite{ye2019cross}\cite{chen2019see}, overall Intersection-over-Union (Overall IoU) and Prec@X are adopted as metrics to evaluate our model. 
Overall IoU calculates total intersection regions over total union regions of all the test samples. 
Prec@X measures the percentage of predictions whose IoU are higher than the threshold $X$ with $X \in \{0.5, 0.6, 0.7, 0.8, 0.9\}$.

\begin{table*}[t]
   \centering
   \begin{tabular}{l|cccc|ccccc|c}
       \hline
        & EP & RAR & TGFE & GloVe & Prec@0.5 & Prec@0.6 & Prec@0.7 & Prec@0.8  & Prec@0.9  & Overall IoU \\
       \hline
       1& & & & &  48.01 & 37.98 & 27.92 & 16.30 & 3.72 & 47.36 \\
      2 & $\surd$ & & & & 49.76 & 40.35 & 30.15 & 17.84 & 4.16 & 49.06 \\
       3 & & $\surd$ & & & 59.32 & 51.16 & 40.59 & 26.50 & 6.66 & 53.40 \\
       4 & $\surd$ & $\surd$ & & & 62.86 & 54.54 & 44.10 & \textbf{28.65} & \textbf{7.24} & 55.38 \\
       5 & $\surd$ & $\surd$ & & $\surd$ & \textbf{62.87} & \textbf{54.91} & \textbf{44.16} & 28.43 & 7.23 & \textbf{56.00} \\
      \hline
      6* & & &  & & 63.12 & 54.56 & 44.20 & 28.75 & 8.51 & 56.38 \\
       7 & & & $\surd$ & & 67.63 & 59.80 & 49.72 & 34.45 & 10.62 & 58.81 \\
       8 & $\surd$ & & $\surd$ & & 68.39 & 60.92 & 50.70 & 35.24 & 11.13 & 59.05 \\
      9 & & $\surd$ & $\surd$ & & 69.37 & 62.28 & 52.66 & 36.89 & 11.27 & 59.62 \\
        10 & $\surd$ & $\surd$ & $\surd$ & & 71.04 & 64.02 & 54.25 & 38.45 & 11.99 & 60.72 \\
        11 & $\surd$ & $\surd$ & $\surd$ & $\surd$ & \textbf{71.27} & \textbf{64.44} & \textbf{55.03} & \textbf{39.28} & \textbf{12.89} & \textbf{61.19} \\
       \hline
   \end{tabular}
   \caption{Ablation studies on UNC val set. *Row $6$ is the multi-level version of row $1$ using only ConvLSTM for fusion. 
   EP and RAR indicate entity perception stage and relation-aware reasoning stage in our CMPC module respectively.}
   \label{tab:cmpc}
\end{table*}

\subsection{Comparison with State-of-the-arts}
To demonstrate the superiority of our method, we evaluate it on four referring segmentation benchmarks. 
Comparison results are presented in Table~\ref{tab:sota}. 
We follow prior works~\cite{ye2019cross}\cite{chen2019see} to only report overall IoU due to the limit of pages. 
Full results are included in supplementary materials. 
As illustrated in Table~\ref{tab:sota}, our method outperforms all the previous state-of-the-arts on 
four benchmarks with large margins. 
Comparing with STEP~\cite{chen2019see} which densely fuses $5$ levels of features for $25$ times, our method 
exploits fewer levels of features and fusion times while consistently achieving $1.40\%$-$2.82\%$ performance gains 
on all the four datasets, demonstrating the effectiveness of our modules. 
In particular, our method yields $2.65\%$ IoU boost against STEP on G-Ref val set, indicating that our method could better handle long sentences than those lack the ability of progressive comprehension. 
Besides, ReferIt is a challenging dataset and previous methods only have marginal improvements on it. For example, 
STEP and CMSA~\cite{ye2019cross} obtain only $0.33\%$ and $0.17\%$ improvements on 
ReferIt test set respectively, while our method enlarges the performance gain to $1.40\%$, 
which shows that our model can well generalize to multiple datasets with different characteristics. 
In addition, our method also outperforms MAttNet~\cite{yu2018mattnet} by a large margin in Overall IoU. 
Though MAttNet achieves higher precisions (e.g., $75.16\%$ versus $71.72\%$ in Prec@0.5 on UNC val set) than ours, it relies on Mask R-CNN~\cite{he2017mask} pretrained on noticeably more COCO~\cite{lin2014microsoft} images ($110K$) than ours pretrained on PASCAL-VOC~\cite{everingham2010pascal} images ($10K$). Therefore, it may not be completely fair to directly compare performances of MAttNet with ours.

\subsection{Ablation Studies}
We perform ablation studies on UNC val set and G-Ref val set to testify the effectiveness of each proposed module.

\textbf{Components of CMPC Module. }We first explore the effectiveness of each component of our proposed CMPC module and the experimental results are shown in Table~\ref{tab:cmpc}.
EP and RAR denotes the entity perception stage and relation-aware reasoning stage in CMPC module respectively. 
GloVe means using GloVe word embeddings~\cite{pennington2014glove} to initialize the embedding layer, 
which is also adopted in~\cite{chen2019see}. 
Results of rows $1$ to $5$ are all based on single-level features, i.e. Res$5$.
Our baseline is implemented as simply concatenating the visual feature extracted with DeepLab-$101$ and linguistic feature extracted with an LSTM and making prediction on the fusion of them.
As shown in row $2$ of Table~\ref{tab:cmpc}, including EP brings $1.70\%$ IoU improvement over the baseline, indicating the perception of candidate entities are essential to the feature alignment between visual and linguistic modalities.
In row $3$, RAR alone brings $6.04\%$ IoU improvement over baseline, which demonstrates that leveraging relational words as routers to reason among spatial regions could effectively highlight the referent in the image, thus boosting the performance notably.
Combining EP with RAR, as shown in row $4$, our CMPC module could achieve $55.38\%$ IoU with single level features, outperforming baseline with a large margin of $8.02\%$ IoU. 
This indicates that our model could accurately identity the referent by progressively comprehending the expression and image. 
Integrated with GloVe word embeddings, the IoU gain further achieves $8.64\%$ with the aid of large-scale corpus.

We further conduct ablation studies based on multi-level features in rows $6$ to $11$ of Table~\ref{tab:cmpc}. 
Row $6$ is the multi-level version of row $1$ using ConvLSTM to fuse multi-level features. 
The TGFE module in rows $7$ to $11$ is based on single round of feature exchange. 
As shown in Table~\ref{tab:cmpc}, our model performs consistently with the single level version, which well proves the effectiveness of our CMPC module.

\begin{figure*}[t]
   \begin{center}
      \includegraphics[width=0.8\linewidth]{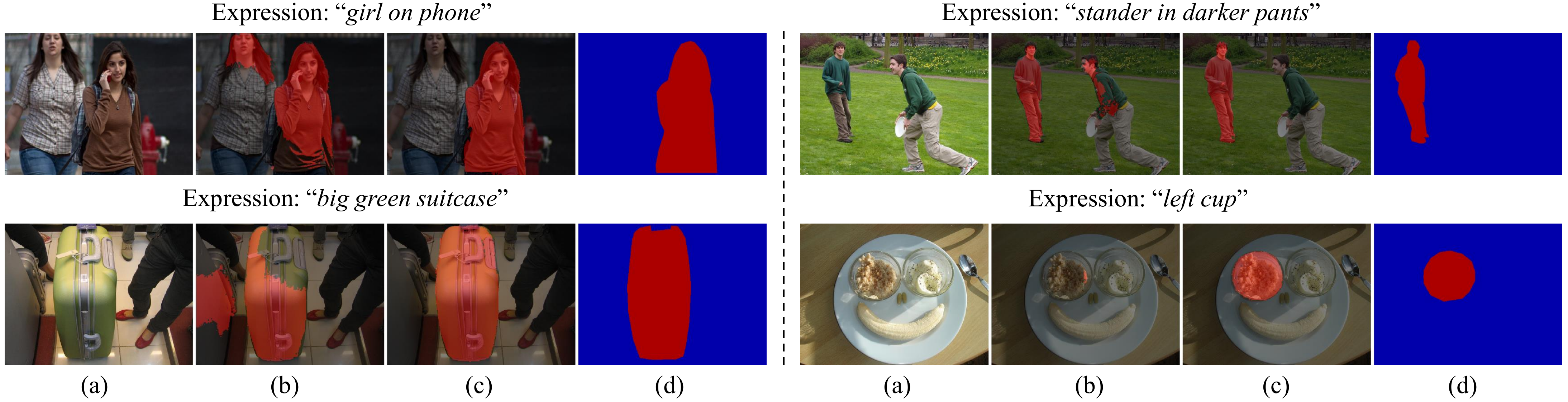}
   \end{center}
      \caption{Qualitative results of referring image segmentation. (a) Original image. (b) Results predicted by the 
      multi-level baseline model (row 6 in Table~\ref{tab:cmpc}). (c) Results predicted by our full model 
      (row 11 in Table~\ref{tab:cmpc}). (d) Ground-truth.}
   \label{fig:qualitative}
\end{figure*}

\begin{figure*}[htbp]
   \begin{center}
      \includegraphics[width=0.7\linewidth]{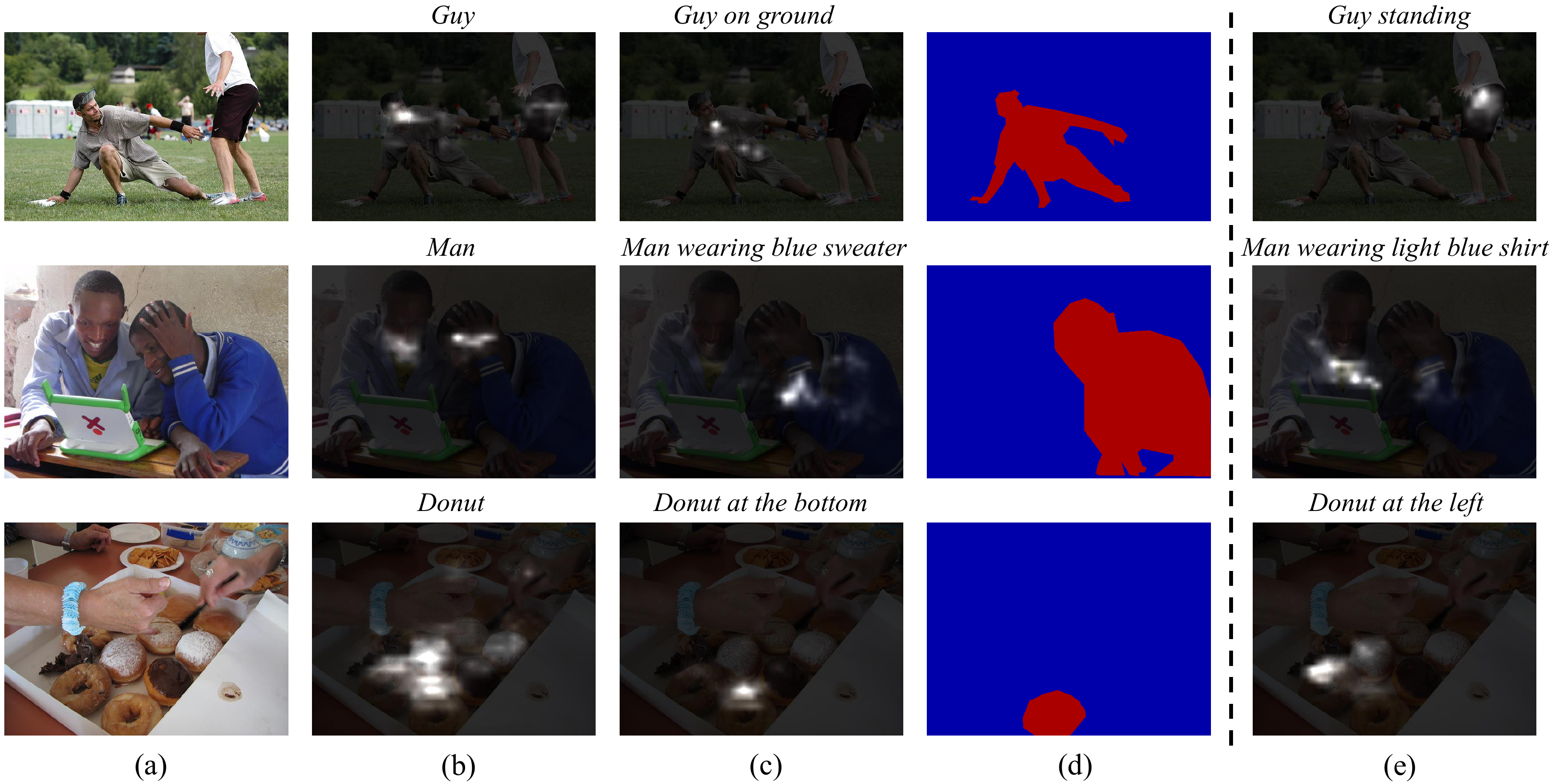}
   \end{center}
      \caption{Visualization of affinity maps between images and expressions in our model. (a) Original image. (b)(c) Affinity maps of only entity words and full expressions in the test samples. (d) Ground-truth. (e) Affinity maps of expressions manually modified by us.}
   \label{fig:attn_map}
\end{figure*}

\textbf{TGFE module. }Table~\ref{tab:tgfe} presents the ablation results of TGFE module. $n$ is the number of feature exchange rounds. The experiments are based on multi-level features with CMPC module. 
It is shown that only one round of feature exchange in TGFE could improve the IoU from $59.85\%$ to $60.72\%$.
When we increase the rounds of feature exchange in TGFE, the IoU increases as well, which well proves the effectiveness of our TGFE module.
We further evaluate TGFE module on baseline model and the comparing results are shown in row $6$ and row $7$ of Table~\ref{tab:cmpc}.
TGFE with single round of feature exchange improves the IoU from $56.38\%$ to $58.81\%$, indicating that our TGFE module can effectively utilize rich contexts in multi-level features.

\vspace{-2mm}
\begin{table}[!htbp]
   \centering
   \begin{tabular}{|c|c|c|c|}
      \hline
      \multirow{2}*{CMPC only} & \multicolumn{3}{|c|}{+TGFE} \\
      \cline{2-4}
      & $n=1$ & $n=2$ & $n=3$ \\
      \hline
      59.85 & 60.72 & 61.07 & \textbf{61.25} \\
      \hline
   \end{tabular}
   \caption{Overall IoUs of different numbers of feature exchange rounds in TGFE module on UNC val set. $n$ denotes the number of feature exchange rounds.}
   \label{tab:tgfe}
\end{table}
\vspace{-6mm}

\begin{table}[!htbp]
   \centering
   \begin{tabular}{|c|c|c|c|c|}
         \hline
         \multirow{2}*{Dataset}  & \multicolumn{4}{|c|}{CMPC} \\
         \cline{2-5}
         & $n=0$ & $n=1$ & $n=2$ & $n=3$ \\
         \hline
         UNC val & 49.06 & \textbf{55.38} & 51.57 & 50.70 \\
         \hline
         G-Ref val & 36.50 & 38.19 & \textbf{40.12} & 38.96 \\
         \hline
   \end{tabular}
   \caption{Experiments of graph convolution on UNC val set and G-Ref val set in terms of \textit{overall IoU}. 
   $n$ denotes the number of graph convolution layers in our CMPC module. 
   Experiments are all conducted on single level features.}
   \label{tab:gcn}
\end{table}

\vspace{-3mm}
\textbf{Number of Graph Convolution Layer. }In Table~\ref{tab:gcn}, we explore the number of graph convolution layers in CMPC module based on single-level features. $n$ is the number of graph convolution layers in CMPC.
Results on UNC val set show that more graph convolution layers leads to performance degradation.
However, on G-Ref val set, $2$ layers of graph convolution in CMPC achieves better performance than $1$ layer while $3$ layers decreasing the performance. 
As the average length of expressions in G-Ref ($8.4$ words) is much longer than that of UNC ($<4$ words), we suppose that stacking more graph convolution layers in CMPC can appropriately improve the reasoning effect for longer referring expressions. 
However, too many graph convolution layers may introduce noises and harm the performance.

\textbf{Qualitative Results. }We presents qualitative comparison between the multi-level baseline model and our full model 
in Figure~\ref{fig:qualitative}. From the top-left example we can observe that the baseline model fails to make clear judgement 
between the two girls, while our full model is able to distinguish the correct girl having relationship with the phone, 
indicating the effectiveness of our CMPC module. Similar result is shown in the top-right example of 
Figure~\ref{fig:qualitative}. As illustrated in the bottom row of Figure~\ref{fig:qualitative}, attributes and location 
relationship can also be well handled by our full model.

\textbf{Visualization of Affinity Maps. }We visualize the affinity maps between multimodal feature and the first word in the expression in Figure~\ref{fig:attn_map}. As shown in (b) and (c), our model is able to progressively produce more concentrated responses on the referent as the expression becomes more informative from only entity words to the full sentence. Interestingly, when we manually modify the expression to refer to other entities in the image, our model is still able to correctly comprehend the new expression and identify the referent. 
For example, in the third row of Figure~\ref{fig:attn_map}(e), when the expression changes from ``Donut at the bottom'' to ``Donut at the left'', high response area shifts from bottom donut to the left donut according to the expression.
It indicates that our model can adapt to new expressions flexibly.

\vspace{-1mm}
\section{Conclusion and Future Work}
\vspace{-1mm}

To address the referring image segmentation problem, we propose a Cross-Modal Progressive Comprehension (CMPC) module which first perceives candidate entities considered by the expression using entity and attribute words, then conduct graph-based reasoning with the aid of relational words to further highlight the referent while suppressing others. 
We also propose a Text-Guided Feature Exchange (TGFE) module which exploits textual information to selectively integrate features from multiple levels to refine the mask prediction. 
Our model consistently outperforms previous state-of-the-art methods on four benchmarks, demonstrating its effectiveness. 
In the future, we plan to analyze the linguistic information more structurally and explore more compact graph formulation.

\vspace{-5mm}
\paragraph{Acknowledgement} This work was partially supported by the National Natural Science Foundation of China (Grant 61572493, Grant 61876177, Grant 61976250, Grant 61702565), Beijing Natural Science Foundation (L182013, 4202034), Fundamental Research Funds for the Central Universities and Zhejiang Lab (No. 2019KD0AB04).

{\small
\bibliographystyle{ieee_fullname}
\bibliography{2428_bib}
}

\end{document}